\pdfoutput=1

\documentclass[11pt]{article}

\usepackage[preprint]{acl}

\usepackage{times}
\usepackage{latexsym}

\usepackage[T1,T2A]{fontenc}

\usepackage[utf8]{inputenc}

\usepackage[russian,english]{babel}

\DeclareFontFamilySubstitution{T2A}{\familydefault}{Tempora-TLF}

\usepackage{microtype}

\usepackage{inconsolata}

\usepackage{graphicx}
\usepackage{xurl} 

\title{The first open machine translation system for the Chechen language}

\author{Abu-Viskhan A. Umishov \\
  \texttt{abuviskhanumishov@gmail.com} \\\And
   Vladislav A. Grigorian\\
  \texttt{real.vladislav.grigorian@yandex.ru} \\
  \AND
  Institute of Mathematics, Mechanics and Computer Sciences, Southern Federal University,\\
Rostov-on-Don, 344090 Russia
}

\begin{document}
\maketitle
\begin{abstract}
We introduce the first open-source model for translation between the vulnerable Chechen language and Russian, and the dataset collected to train and evaluate it. We explore fine-tuning capabilities for including a new language into a large language model system for multilingual translation NLLB-200. The BLEU\;/\;ChrF++ scores for our model are 8.34\;/\;34.69 and 20.89\;/\;44.55 for translation from Russian to Chechen and reverse direction, respectively. The release of the translation models is accompanied by the distribution of parallel words, phrases and sentences corpora and multilingual sentence encoder adapted to the Chechen language.
\end{abstract}

\section{Introduction}

In the Russian Federation alone, Chechen is spoken by approximately 1.5 million people, and more than 97 percent of them use it in everyday life \cite{rosstat_chechen_language}. Chechen was added to Google Translate\footnote{\url{https://blog.google/products/translate/google-translate-new-languages-2024/}} in 2024, but to our knowledge no open translation systems for Chechen have been published, although there is a baseline for the English-Chechen pair \cite{kudugunta2023madlad400multilingualdocumentlevellarge}. Our work is inspired by \citealp{dale2022neuralmachinetranslationerzya}, whose author has created a translator for the previously uncovered Erzyan based on the mBART50 model \cite{tang2020multilingualtranslationextensiblemultilingual}. Like the author of the paper, we had only publicly available data to build a parallel corpus and a very small budget, but we chose the more modern nllb-200-distilled-600M\footnote{\url{https://huggingface.co/facebook/nllb-200-distilled-600M}} model \cite{nllbteam2022languageleftbehindscaling} as the basis for our model.

Chechen belongs to the Nakh branch of the Nakh-Dagestani language family, together with Ingush and Batsbi. Chechen is one of the state languages of the Chechen Republic and Dagestan, and is widely spoken in Ingushetia and the rest of southern Russia. There are native-speaking Chechen diasporas in many European, Middle Eastern, and Central Asian countries. Ingush, a close relative of Chechen, is one of the state languages of the Republic of Ingushetia, and Batsbi is spoken in the Tusheti region of Georgia. The written Chechen language is now based on the Cyrillic alphabet.

Books and magazines are published in Chechen, 24-hour television and radio are broadcast, and keyboard layouts for the Cyrillic script have been developed , however, the state of the language is considered vulnerable (VU according to UNESCO classification) \footnote{\url{https://en.wal.unesco.org/countries/russian-federation/languages/chechen}}.

As part of our work, we present and publish the following results prepared for Chechen:
\begin{itemize}
    \item      A sentence encoder model based on LaBSE \cite{feng2022languageagnosticbertsentenceembedding}\footnote{\url{https://huggingface.co/NM-development/LaBSE-en-ru-ce-prototype}}
    \item Chechen-Russian parallel corpus\footnote{\url{https://huggingface.co/datasets/NM-development/nmd-ce-ru-171k-v0}}
    \item Neural translation model for translation between Chechen and Russian language based on nllb-200-distilled-600M \cite{nllbteam2022languageleftbehindscaling}\footnote{\url{https://huggingface.co/NM-development/nllb-ce-rus-v0}}
\end{itemize}

We evaluated the quality of our translation model's performance between Chechen, Russian and English, using the BLEU and ChrF++ metrics. We conducted a human assessment on a five-point scale for translations between Chechen and Russian. In all cases, we compared our model's quality metrics to existing translation solutions to assess its proficiency.

Our evaluation shows that all open-source translation models known to us, for which it is claimed to be able to work with Chechen, produce extremely poor translations in the most demanded directions, i.e. ru-ce, ce-ru, en-ce, ce-en. In turn, the translation quality of our model in ru-ce and ce-ru is close to that of Google Translator and Claude 3.7 Sonnet.

\section{Related Work}

There are multiple published monolingual and parallel corpus for Chechen language, namely, Chechen Text Corpus\footnote{\url{https://baltoslav.eu/nox/}}, Chechen language corpus collected from open sources\footnote{\url{https://corpora.dosham.info/}},  MADLAD-400 \cite{kudugunta2023madlad400multilingualdocumentlevellarge} and Gatios \cite{jones2023bilexrxlexicaldata} datasets.

MADLAD-400 \cite{kudugunta2023madlad400multilingualdocumentlevellarge} and Helsinki-NLP community's\footnote{\url{https://huggingface.co/Helsinki-NLP/opus-mt-mul-en}, \url{https://huggingface.co/Helsinki-NLP/opus-mt-en-mul}} models provided a baseline for Chechen translation, however during the evaluation process, we show that they are not very useful for translation tasks.

nllb-200-distilled-600M \cite{nllbteam2022languageleftbehindscaling} has become the most popular open-source machine translation model, especially for "low-resource" languages\footnote{\url{https://huggingface.co/models?pipeline_tag=translation}}. In the work \citealp{dale2022neuralmachinetranslationerzya}, it was shown that the model successfully learns languages that were not present in the original parallel corpus, using Erzyan as an example. In addition, it has been shown that the model can be trained on languages from families that were not present in the original parallel corpora: although the original training dataset did not include any languages from the Nakh-Dagestani family, recent experiments have shown that the model can be successfully trained on these languages using transfer learning methods, using Lezgian as an example \cite{Asvarov_2024}.

\section{Methodology}
\subsection{Data collection}
We used data from various sources:
\begin{itemize}
    \item Parallel sentences from the Bible \footnote{\url{https://ibt.org.ru/chechenskiy/vsya-bibliya/chitat}}
    \item Parallel sentences from merged from two Chechen translations of the Quran by Magomed Magomedov and Adam Ibragimov and three Russian translations by Elmir Quliyev, Abu Adel and Magomed-Nuri Osmanov \footnote{\url{https://ru.quranacademy.org/quran}}
    \item Parallel words and short phrases from four dictionaries
    \begin{itemize}
        \item Chechen-Russian Dictionary A.T. Karasaev, A.G. Matsiev \footnote{\url{https://dosham.wordpress.com/}}
        \item Chechen-Russian, Russian-Chechen dictionary of human anatomy Bersanov R.U. \footnote{\url{https://ps95.ru/ce/download/848/}}
        \item Russian-Chechen, Chechen-Russian Dictionary of computer vocabulary S.M. Umarkhadzhiev, A.V. Astemirov, H.I. Askhabov, A.S. Badaeva, A.D. Vagapov, E.S. Izrailova, Z.A. Sultanov \footnote{\url{https://ps95.ru/download/731/}}
        \item BaltoSlav Short Chechen-Russian Dictionary \footnote{\url{https://baltoslav.eu/noxciyn/slounik.php?mova=ru}}
    \end{itemize}
    \item Gatios dataset for Chechen-English parallel words and short phrases  \cite{jones2023bilexrxlexicaldata}
    \item 3 books with parallel Chechen and Russian translation \footnote{\url{https://rus4all.ru/che/}, \url{https://vayvault.com/}}
    \item News articles in Chechen from Daimohk website \footnote{\url{https://daymohk-gazet.ru/}}
    \item Numbers generated in Chechen and Russian with num2words Python library \footnote{\url{https://github.com/savoirfairelinux/num2words}}
\end{itemize}

We parsed and normalized Chechen and Russian texts, collected pairs of sentences with markup, and aligned those texts that did not have markup with our sentence encoder.\\
In total, 171K Chechen-Russian sentence pairs, and 481K monolingual Chechen sentences were obtained. For the parallel corpus, the distribution of its sources is shown in Table \ref{tab:sources_distribution}.

\begin{table}[h]
\addtolength{\tabcolsep}{-2pt} 
  \centering
  \begin{tabular}{l|c}
    \hline
    \textbf{Source} & \textbf{Proportion, \%} \\
    \hline
    \verb|Dictionaries| & {57}\\
    \verb|Quran|     & {22} \\
    \verb|Bible|     & {17}  \\
    \verb|Gatitos|     & {3}\\
    \verb|Numbers (num2words)| & {1} \\
    \verb|News and fiction| & {1} \\\hline
  \end{tabular}
  \caption{The distribution of sources across the parallel corpus}
  \label{tab:sources_distribution}
\end{table}
\newpage

We also estimate the proportion of words and short phrases, on the one hand, and sentences, on the other, in our dataset. We estimate this proportion in three different ways: as the proportion of rows corresponding to each category, as the proportion of words, and as the proportion of symbols representing each category. These proportions are calculated for both the Chechen and Russian languages, and they are shown in Appendices \ref{app:length_proportions_train} and \ref{app:length_proportions_eval}, respectively, for the training and evaluation datasets.

\subsection{Data augmentation}
We joined together consecutive parallel sentence pairs and triplets from news text and appended resulting text pairs into our training dataset. 

\subsection{Chechen Sentence Encoder}
To compute sentence embeddings, we use an encoder based on LaBSE \cite{feng2022languageagnosticbertsentenceembedding} and followed the approach described in \citealp{dale2022neuralmachinetranslationerzya}. We used the model's version truncated for English and Russian languages \footnote{\url{https://huggingface.co/cointegrated/LaBSE-en-ru}}. We added new tokens for Chechen language by training BPE \cite{sennrich2016neuralmachinetranslationrare} tokenizer over monolingual corpus and fine-tuned it with Chechen-Russian parallel data.

We used this trained sentence encoder during parallel corpus collection to align unlabelled texts into Chechen-Russian sentence pairs.

\subsection{Training Machine Translation Models}
We extended the SentencePiece \cite{kudo2018sentencepiecesimplelanguageindependent} vocabulary of the nllb-200-distilled-600 model \cite{nllbteam2022languageleftbehindscaling} with \textit{ce\_Cyrl} special token and approximately 16K Chechen tokens using methods described in David Dale's blog post\footnote{\url{https://cointegrated.medium.com/how-to-fine-tune-a-nllb-200-model-for-translating-a-new-language-a37fc706b865}}.\\
The embeddings for the new tokens are initialized as the averages of the embeddings of their subtokens, inspired by \citealp{xu2022subwordalignmentusefulvestpocket}.\\
Training hyperparameters are presented in Appendix \ref{app:hyperparams}.


\section{Evaluation}

\subsection{Data and inference parameters}
We evaluate our translation model on a holdout dataset of size 360, taken from the shuffled corpus. We excluded sources with sentences containing multiple parallel paras from the evaluation dataset to ensure that the model wasn't trained on similar data.\\
We expanded our original Chechen-Russian benchmark for English language by translating its Russian part into English with Google Translator.\\
The inference parameters of our translation model are presented in Appendix \ref{app:inference_params}.

\subsection{Automated Metrics}

For both evaluated directions we calculate BLEU \cite{papineni-etal-2002-bleu, post2018clarityreportingbleuscores} and ChrF++ \cite{popovic-2017-chrf} for our model and compared them to existing translation models that have been stated to be capable of working with Chechen and Claude 3.7 Sonnet model for text generation with special prompt for translation task taken from \citealp{mamasaidov2024openlanguagedatainitiative} (see Appendix \ref{app:trans_prompt}). The values of these metrics on the evaluation set are given in Tables \ref{tab:bleu_eval_table} and \ref{tab:chrf_eval_table}. As there was no direct translation option between Chechen in Russian in Helsinki-NLP community's models we combined together Helsinki-NLP/opus-mt-mul-en\footnote{\url{https://huggingface.co/Helsinki-NLP/opus-mt-mul-en}} and  Helsinki-NLP/opus-mt-en-mul\footnote{\url{https://huggingface.co/Helsinki-NLP/opus-mt-en-mul}} models for this purpose.\\
We also calculate these metrics on subsets of different sources of our evaluation dataset and provide the resulting BLEU\;/\;ChrF++ scores in Table \ref{tab:source_eval_table}.\\
A significant discrepancy in quality was identified between translation directions. The underlying causes of this variation remain to be elucidated.

\begin{table}[h]
\addtolength{\tabcolsep}{-2pt} 
  \centering
  \begin{tabular}{l|cccc}
    \hline
    \textbf{Model} & \textbf{ce-ru} & \textbf{ru-ce} & \textbf{ce-en} & \textbf{en-ce}  \\
    \hline
    \verb|NLLB-200|     & {20.89}  & {8.34} & {0.09} & {4.38}        \\
    \verb|Helsinki-NLP|     & {0.41}  & {0.06} & {0.56} & {0.04}      \\
    \verb|MADLAD-400|     & {0.34}  & {0.30} & {0.42} & {0.05}         \\
    \verb|Google Translate|     & {26.83}  & {11.74} & {26.78} & {11.48}    \\
    \verb|Claude 3.7 Sonnet|     & {14.28}  & {5.80} & {17.98} & {4.06}  \\\hline
  \end{tabular}
  \caption{BLEU scores for several models on the evaluation set}
  \label{tab:bleu_eval_table}
\end{table}

\begin{table}[h]
\addtolength{\tabcolsep}{-2pt} 
  \centering
  \begin{tabular}{l|cccc}
    \hline
    \textbf{Model} & $\textbf{ce-ru}$ & \textbf{ru-ce} & \textbf{ce-en} & \textbf{en-ce}  \\
    \hline
    \verb|NLLB-200|     & {44.55}  & {34.69} & {1.57} & {23.15}        \\
    \verb|Helsinki-NLP|     & {11.29}  & {1.23} & {13.26} & {1.24}      \\
    \verb|MADLAD-400|     & {11.31}  & {3.96} & {11.79} & {3.55}         \\
    \verb|Google Translate|     & {44.03}  & {37.16} & {46.89} & {36.65}      \\
    \verb|Claude 3.7 Sonnet|     & {38.22}  & {30.56} & {41.87} & {28.57}  \\\hline
  \end{tabular}
  \caption{ChrF++ scores for several models on the evaluation set}
  \label{tab:chrf_eval_table}
\end{table}

\begin{table}[h!]
\addtolength{\tabcolsep}{-2pt} 
  \centering
  \begin{tabular}{l|cc}
    \hline
    \textbf{Source} & \textbf{ce-ru} & \textbf{ru-ce}   \\
    \hline
    \verb|All|    & {20.89\;/\;44.55}  & {8.34\;/\;34.69}  \\
    \verb|Bible|    & {20.22\;/\;45.01} & {8.32\;/\;33.49}  \\
    \verb|Dictionaries|  & {23.12\;/\;41.21} & {9.93\;/\;36.59} \\
    \verb|Fiction|    & {8.12\;/\;30.68} & {1.74\;/\;29.74}  \\
    \verb|Gatitos|    & {52.40\;/\;50.92} & {18.15\;/\;35.81}  \\
    \verb|News|    & {22.36\;/\;44.78} & {7.35\;/\;39.06}     \\
    \verb|Numbers|    & {100.00\;/\;100.00} & {26.18\;/\;69.07}   \\\hline
  \end{tabular}
  \caption{BLEU\;/\;ChrF++ scores for different sources}
  \label{tab:source_eval_table}
\end{table}

\subsection{Manual Evaluation}\label{sec:manual_eval}
To evaluate our translation model and compare it with closed-source solutions that have demonstrated acceptable quality according to automated metrics, we asked four native Chechen speakers to assess the quality of the translation on the evaluation dataset.
For the evaluation process we used the protocol described in \citealp{dale2022neuralmachinetranslationerzya}, based on the XSTS protocol \cite{nllbteam2022languageleftbehindscaling}. The scores are between 1 (a useless translation) and 5 (a perfect translation), with 3 points standing for an acceptable translation without serious errors.\\
For each participant in the experiment, we randomly selected sixty translated sentences from the evaluation dataset. Twenty of these sentences were translated in both directions by our model, twenty were translated by Google Translate, and the remaining twenty were translated using Claude 3.7 Sonnet. The volunteers were asked to rate the translations from Chechen to Russian and vice versa. To ensure the impartiality of the assessment, the model used for each translation was kept hidden.\\
The average estimation for each model is presented in Table \ref{tab:human_eval_table}. The average proportion of translations that were deemed acceptable (human evaluation score $\geq3$) is shown in Table \ref{tab:acceptance_rate_table}.

\begin{table}[h!]
\addtolength{\tabcolsep}{-2pt} 
  \centering
  \begin{tabular}{l|cc}
    \hline
    \textbf{Model} & \textbf{ce-ru} & \textbf{ru-ce} \\
    \hline
    \verb|NLLB-200|     & {3.9}  & {3.9}    \\
    \verb|Google Translate|     & {3.2}  & {3.6}    \\
    \verb|Claude 3.7 Sonnet|     & {3.5}  & {3.6} \\\hline
  \end{tabular}
  \caption{Human evaluation scores for several models on the evaluation set}
  \label{tab:human_eval_table}
\end{table}

\begin{table}[h!]
\addtolength{\tabcolsep}{-2pt} 
  \centering
  \begin{tabular}{l|cc}
    \hline
    \textbf{Model} & \textbf{ce-ru} & \textbf{ru-ce} \\
    \hline
    \verb|NLLB-200|     & {84}  & {85}    \\
    \verb|Google Translate|     & {69}  & {84}    \\
    \verb|Claude 3.7 Sonnet|     & {76}  & {81} \\\hline
  \end{tabular}
  \caption{The proportion of acceptable translations in percent for several models on the evaluation set}
  \label{tab:acceptance_rate_table}
\end{table}
\section{Limitations}
The majority of the training data for our translation model was comprised of parallel corpora consisting of word pairs, short phrases, and sentences. As the behavior of the model within a longer context has not yet been investigated, it may be difficult to predict.
\section{Conclusion}
Our work shows that using data available on the Internet with small computing power, it is possible to teach the Chechen language an existing translation model and obtain quality comparable to closed-source solutions.\\
In the course of our work, we have collected 171,000 parallel Chechen-Russian sentences. We have also trained an open-source translation system for the Chechen and Russian languages, as well as a BERT-based sentence encoder for the Chechen language. These resources are all publicly available.\\
We hope that the data we share and the fine-tuned models we publish will contribute to the development of various language models within the community of Chechen language enthusiasts and businesses focusing on the Chechen language.

\section{Acknowledgments}
We are grateful to David Dale for the advice and review of this article, for his blog on Medium and Habr, and for the efforts he makes to ensure that the low-resource languages of the world have their own translation models.

\newpage 
\bibliography{custom}

\begin{thebibliography}{21}
\providecommand{\natexlab}[1]{#1}

\bibitem[{rus()}]{rus4all}

\newblock \href {https://rus4all.ru/che/} {Портал национальных литератур. Чеченский язык.}

\bibitem[{Asvarov and Grabovoy(2024)}]{Asvarov_2024}
Alidar Asvarov and Andrey Grabovoy. 2024.
\newblock \href {https://doi.org/10.1109/ispras64596.2024.10899143} {Neural machine translation system for lezgian, russian and azerbaijani languages}.
\newblock In \emph{2024 Ivannikov Ispras Open Conference (ISPRAS)}, page 1–7. IEEE.

\bibitem[{BaltoSlav()}]{baltoslav}
BaltoSlav.
\newblock \href {https://baltoslav.eu/nox/} {Chechen text corpus}.

\bibitem[{Dale(2022)}]{dale2022neuralmachinetranslationerzya}
David Dale. 2022.
\newblock \href {https://arxiv.org/abs/2209.09368} {The first neural machine translation system for the erzya language}.
\newblock \emph{Preprint}, arXiv:2209.09368.

\bibitem[{Feng et~al.(2022)Feng, Yang, Cer, Arivazhagan, and Wang}]{feng2022languageagnosticbertsentenceembedding}
Fangxiaoyu Feng, Yinfei Yang, Daniel Cer, Naveen Arivazhagan, and Wei Wang. 2022.
\newblock \href {https://arxiv.org/abs/2007.01852} {Language-agnostic bert sentence embedding}.
\newblock \emph{Preprint}, arXiv:2007.01852.

\bibitem[{Jones et~al.(2023)Jones, Caswell, Saxena, and Firat}]{jones2023bilexrxlexicaldata}
Alex Jones, Isaac Caswell, Ishank Saxena, and Orhan Firat. 2023.
\newblock \href {https://arxiv.org/abs/2303.15265} {Bilex rx: Lexical data augmentation for massively multilingual machine translation}.
\newblock \emph{Preprint}, arXiv:2303.15265.

\bibitem[{Khasbulatov()}]{corporadosh}
Yusuf Khasbulatov.
\newblock \href {https://corpora.dosham.info/} {Chechen language corpus collected from open sources}.

\bibitem[{Kudo and Richardson(2018)}]{kudo2018sentencepiecesimplelanguageindependent}
Taku Kudo and John Richardson. 2018.
\newblock \href {https://arxiv.org/abs/1808.06226} {Sentencepiece: A simple and language independent subword tokenizer and detokenizer for neural text processing}.
\newblock \emph{Preprint}, arXiv:1808.06226.

\bibitem[{Kudugunta et~al.(2023)Kudugunta, Caswell, Zhang, Garcia, Choquette-Choo, Lee, Xin, Kusupati, Stella, Bapna, and Firat}]{kudugunta2023madlad400multilingualdocumentlevellarge}
Sneha Kudugunta, Isaac Caswell, Biao Zhang, Xavier Garcia, Christopher~A. Choquette-Choo, Katherine Lee, Derrick Xin, Aditya Kusupati, Romi Stella, Ankur Bapna, and Orhan Firat. 2023.
\newblock \href {https://arxiv.org/abs/2309.04662} {Madlad-400: A multilingual and document-level large audited dataset}.
\newblock \emph{Preprint}, arXiv:2309.04662.

\bibitem[{Mamasaidov and Shopulatov(2024)}]{mamasaidov2024openlanguagedatainitiative}
Mukhammadsaid Mamasaidov and Abror Shopulatov. 2024.
\newblock \href {https://arxiv.org/abs/2409.04269} {Open language data initiative: Advancing low-resource machine translation for karakalpak}.
\newblock \emph{Preprint}, arXiv:2409.04269.

\bibitem[{Papineni et~al.(2002)Papineni, Roukos, Ward, and Zhu}]{papineni-etal-2002-bleu}
Kishore Papineni, Salim Roukos, Todd Ward, and Wei-Jing Zhu. 2002.
\newblock \href {https://doi.org/10.3115/1073083.1073135} {{B}leu: a method for automatic evaluation of machine translation}.
\newblock In \emph{Proceedings of the 40th Annual Meeting of the Association for Computational Linguistics}, pages 311--318, Philadelphia, Pennsylvania, USA. Association for Computational Linguistics.

\bibitem[{Popovi{\'c}(2017)}]{popovic-2017-chrf}
Maja Popovi{\'c}. 2017.
\newblock \href {https://doi.org/10.18653/v1/W17-4770} {chr{F}++: words helping character n-grams}.
\newblock In \emph{Proceedings of the Second Conference on Machine Translation}, pages 612--618, Copenhagen, Denmark. Association for Computational Linguistics.

\bibitem[{Post(2018)}]{post2018clarityreportingbleuscores}
Matt Post. 2018.
\newblock \href {https://arxiv.org/abs/1804.08771} {A call for clarity in reporting bleu scores}.
\newblock \emph{Preprint}, arXiv:1804.08771.

\bibitem[{Sennrich et~al.(2016)Sennrich, Haddow, and Birch}]{sennrich2016neuralmachinetranslationrare}
Rico Sennrich, Barry Haddow, and Alexandra Birch. 2016.
\newblock \href {https://arxiv.org/abs/1508.07909} {Neural machine translation of rare words with subword units}.
\newblock \emph{Preprint}, arXiv:1508.07909.

\bibitem[{Tang et~al.(2020)Tang, Tran, Li, Chen, Goyal, Chaudhary, Gu, and Fan}]{tang2020multilingualtranslationextensiblemultilingual}
Yuqing Tang, Chau Tran, Xian Li, Peng-Jen Chen, Naman Goyal, Vishrav Chaudhary, Jiatao Gu, and Angela Fan. 2020.
\newblock \href {https://arxiv.org/abs/2008.00401} {Multilingual translation with extensible multilingual pretraining and finetuning}.
\newblock \emph{Preprint}, arXiv:2008.00401.

\bibitem[{Team et~al.(2022)Team, Costa-jussà, Cross, Çelebi, Elbayad, Heafield, Heffernan, Kalbassi, Lam, Licht, Maillard, Sun, Wang, Wenzek, Youngblood, Akula, Barrault, Gonzalez, Hansanti, Hoffman, Jarrett, Sadagopan, Rowe, Spruit, Tran, Andrews, Ayan, Bhosale, Edunov, Fan, Gao, Goswami, Guzmán, Koehn, Mourachko, Ropers, Saleem, Schwenk, and Wang}]{nllbteam2022languageleftbehindscaling}
NLLB Team, Marta~R. Costa-jussà, James Cross, Onur Çelebi, Maha Elbayad, Kenneth Heafield, Kevin Heffernan, Elahe Kalbassi, Janice Lam, Daniel Licht, Jean Maillard, Anna Sun, Skyler Wang, Guillaume Wenzek, Al~Youngblood, Bapi Akula, Loic Barrault, Gabriel~Mejia Gonzalez, Prangthip Hansanti, and 20 others. 2022.
\newblock \href {https://arxiv.org/abs/2207.04672} {No language left behind: Scaling human-centered machine translation}.
\newblock \emph{Preprint}, arXiv:2207.04672.

\bibitem[{Xu and Hong(2022)}]{xu2022subwordalignmentusefulvestpocket}
Minhan Xu and Yu~Hong. 2022.
\newblock \href {https://arxiv.org/abs/2205.04067} {Sub-word alignment is still useful: A vest-pocket method for enhancing low-resource machine translation}.
\newblock \emph{Preprint}, arXiv:2205.04067.

\bibitem[{Берсанов(2010)}]{bersanov_dictionary}
Р.~У. Берсанов. 2010.
\newblock Анатомия человека чеченско-русский атлас, латино-русско-чеченский словарь терминов.

\bibitem[{Карасаев and Мациев(1978)}]{matciev_dictionary}
А.Т. Карасаев and А.Г. Мациев. 1978.
\newblock Русско-чеченский словарь.

\bibitem[{Росстат(2020)}]{rosstat_chechen_language}
Росстат. 2020.
\newblock \href {https://rosstat.gov.ru/vpn/2020/Tom5_Nacionalnyj_sostav_i_vladenie_yazykami} {Итоги ВПН-2020. Том 5 Национальный состав и владение языками}.

\bibitem[{Умархаджиев et~al.(2016)Умархаджиев, Астемиров, Асхабов, Бадаева, Вагапов, Израилова, and Султанов}]{umarhajiev_comp_dictionary}
С.М. Умархаджиев, А.В. Астемиров, Х.И. Асхабов, А.С. Бадаева, А.Д. Вагапов, Э.С. Израилова, and З.А. Султанов. 2016.
\newblock Русско-чеченский, чеченско-русский словарь компьютерной лексики.

\end{thebibliography}
\nocite{*}

\clearpage

\appendix
\onecolumn

\section{The proportions in which words and short phrases relate to sentences in the training dataset}\label{app:length_proportions_train}

\begin{table}[h]
\addtolength{\tabcolsep}{-2pt} 
  \centering
  \begin{tabular}{l|c|c}
     & \textbf{ce} & \textbf{ru} \\
    \hline
    \verb|Rows|     & {61\;/\;39}  & {61\;/\;39}    \\
    \verb|Words|     & {14\;/\;86}  & {13\;/\;87}    \\
    \verb|Symbols|     & {14\;/\;86}  & {16\;/\;84} \\\hline
  \end{tabular}
  \caption{Proportion of words and short phrases \;/\; proportion of sentences, both expressed as a percentage}
  \label{tab:length_proportions_train}
\end{table}

\section{The proportions in which words and short phrases relate to sentences in the evaluation dataset}\label{app:length_proportions_eval}

\begin{table}[h]
\addtolength{\tabcolsep}{-2pt} 
  \centering
  \begin{tabular}{l|c|c}
     & \textbf{ce} & \textbf{ru} \\
    \hline
    \verb|Rows|     & {73\;/\;27}  & {73\;/\;27}    \\
    \verb|Words|     & {20\;/\;80}  & {21\;/\;79}    \\
    \verb|Symbols|     & {21\;/\;79}  & {26\;/\;74} \\\hline
  \end{tabular}
  \caption{Proportion of words and short phrases \;/\; proportion of sentences, both expressed as a percentage}
  \label{tab:length_proportions_eval}
\end{table}

\section{Training hyperparameters}\label{app:hyperparams}

\begin{table}[h]
\centering
\small  
\begin{tabular}{ll}
\textbf{Hyperparameter} & \textbf{Value} \\
\hline
Learning rate & 1e-4 \\
Batch size & 64 \\
Epochs & 9 \\
Optimizer & Adafactor \\
LR scheduler & Constant learning rate with linear warmup \\
Weight decay & 1e-3 \\
Maximum sequence length & 128 \\
Number of warmup steps & 1500
\end{tabular}
\caption{Training hyperparameters}
\label{tab:hyperparams}
\end{table}

\section{Inference parameters}\label{app:inference_params}
\begin{table}[h!]
\small
\begin{tabular}{ll}
\textbf{Parameter} & \textbf{Value} \\
\hline
Temperature & 1.0 \\
Top-k sampling & 50 \\
Top-p sampling & 1.0 \\
Beam search width & 4 \\
Repetition penalty & 1.0 \\
Max output length & 1024 \\
\end{tabular}
\caption{Inference parameters}
\label{tab:inference_params}
\end{table}

\section{Prompt for translation from Chechen to Russian using Claude 3.7 Sonnet}\label{app:trans_prompt}
\begin{verbatim}
You are a professional translator specializing in {source_language} to {target_language} translations.
Your task is to translate the given {source_language} text into {target_language} with the highest
level of accuracy, preserving the original meaning and context. Use proper grammar, punctuation, and
idiomatic expressions appropriate for {target_language} speakers.
Do not include any additional explanations or commentary; provide only the translated text.
{source_language}: {text}
{target_language}:
\end{verbatim}

\clearpage
\section{Quality annotation guidelines}
    \cite{dale2022neuralmachinetranslationerzya} The following annotation criteria (in Russian) were suggested to the annotators in Section \ref{sec:manual_eval}.
\begin{itemize}
    \item 5 points: a perfect translation. The meaning and the style are reproduced completely, the grammar and word choice are correct, the text looks natural.
    \item 4 points: a good translation. The meaning is reproduced completely or almost completely, the style and the word choice are natural for the target language.
    \item 3 points: an acceptable translation. The general meaning is reproduced; the mistakes in word choice and grammar do not hinder understanding; most of the text is grammatically correct and in the target language.
    \item 2 points: a bad translation. The text is mainly understandable and mainly in the target language, but there are critical mistakes in meaning, grammar, or word choice.
    \item 1 point: a useless translation. A large part of the text is in the wrong language, or is incomprehensible, or has little relation to the original text.
\end{itemize}

\section{Translation examples from the evaluation dataset}\label{sec:trans_examples}

\begin{table}[h!]
\scriptsize 
\addtolength{\tabcolsep}{-2pt} 
  \centering
  \begin{tabular}{l|c}
    \hline
    \textbf{Type} & $\textbf{Text}$ \\\hline
    \verb|Source (ce)| & \verb|Кхаьънаш ма эца, хӀунда аьлча цара са гуш верг бӀаьрзе во, бакъболчеран гӀуллакх талхадо.| \\
    \verb|Source (ru)| & \verb|Даров не принимай, ибо дары слепыми делают зрячих и превращают дело правых.| \\
    \verb|Source (en)| & \verb|Do not accept a bribe, for a bribe blinds those who see and twists the words of the innocent.| \\
    \verb|Translation (ce2ru)| & \verb|Не покупай драгоценностей, потому что они ослепляют прозорливца, и дело праведных превращает в труху.| \\
    \verb|Translation (ru2ce)| & \verb|Хьайна луш долугӀат ма къобалде, аьлча хьайна лушгӀаташа бӀаьрсадоцурш а, лаамехь берш а дахьовзор бу.| \\\hline
    \verb|Source (ce)| & \verb|– Вайн тӀеман а, махлелоран а кеманаш йукъаозор ду-кх.| \\
    \verb|Source (ru)| & \verb|— Подключим суда нашего торгового и военного флота.| \\
    \verb|Source (en)| & \verb|— We will use our merchant and naval vessels.| \\
    \verb|Translation (ce2ru)| & \verb|— Объединены наши военные и торговые суда.| \\
    \verb|Translation (ru2ce)| & \verb|Вайн махлелоран а, а флотин а кеманаш вовшахтаса.| \\\hline
    \verb|Source (ce)| & \verb|Хин Ӏоврех хьоьгуш болу сай санна, сан са а ду Хьох хьоьгуш, Дела!| \\
    \verb|Source (ru)| & \verb|Как лань желает к потокам воды, так желает душа моя к Тебе, Боже!| \\
    \verb|Source (en)| & \verb|As the deer longs for streams of water, so I long for you, O God.| \\
    \verb|Translation (ce2ru)| & \verb|Как орел жаждет реки, так жаждет душа моя, Боже!| \\
    \verb|Translation (ru2ce)| & \verb|ХӀунда аьлча сан са Хьоьга,, хьоьжуш ду, хордан хин Ӏоврашка хьоьжуш.| \\\hline
    \verb|Source (ce)| & \verb|ХӀан-хӀа, дайшкара схьа дворянех волчу Михаил Тариэловича ша лахвийр вац цунах хьегарца.| \\
    \verb|Source (ru)| & \verb|Нет, Михаил Тариэлович, потомственный дворянин, не опустится до такого.| \\
    \verb|Source (en)| & \verb|No, the hereditary nobleman Michael Tarielovich won't fall to such a thing.| \\
    \verb|Translation (ce2ru)| & \verb|Нет, Михаил Тариэлович из соседних дворян не одолеет его поступка.| \\
    \verb|Translation (ru2ce)| & \verb|Дера, цу тайпана тайпанара эла, Михаил Тариэлович, кхузахь охьавуссур.| \\\hline
    \verb|Source (ce)| & \verb|Стигална кӀел къахьоьгуш, ша мел динчу хӀуманах буьсун болу хӀун пайда оьцу адамо?| \\
    \verb|Source (ru)| & \verb|Что пользы человеку от всех трудов его, которыми трудится он под солнцем?| \\
    \verb|Source (en)| & \verb|What do people gain from all their labors at which they toil under the sun?| \\
    \verb|Translation (ce2ru)| & \verb|Что пользы человеку от того, что он трудился под солнцем?| \\
    \verb|Translation (ru2ce)| & \verb|Стенах хун пайда бу адамна цо къахьегначу балхах?| \\\hline
    \verb|Source (ce)| & \verb|ТӀеман тӀаьххьарчу шерашкахь коьрта тидам айкхашна тӀеберзийра Евдокимовс.| \\
    \verb|Source (ru)| & \verb|В последние годы войны Евдокимов все больше внимания уделял лазутчикам.| \\
    \verb|Source (en)| & \verb|In the last years of the war, Evdokimov paid more attention to informers.| \\
    \verb|Translation (ce2ru)| & \verb|В последние годы войны Евдокимов обратил внимание на высоты.| \\
    \verb|Translation (ru2ce)| & \verb|Том дабаханчу шерашкахь Евдокимовс алсам тидам бора ладугӀу.| \\\hline
    \verb|Source (ce)| & \verb|– Нохчийчуьра хаамаш буй?| \\
    \verb|Source (ru)| & \verb|— Что сообщают из Чечни?| \\
    \verb|Source (en)| & \verb|Are there any messages from Chechnya?| \\
    \verb|Translation (ce2ru)| & \verb|— Слышены ли из Чечни известия?| \\
    \verb|Translation (ru2ce)| & \verb|ХӀун ду Нохчийчуьра?| \\\hline
  \end{tabular}
  \caption{Translation examples}
  \label{tab:trans_examples}
\end{table}

\end{document}